\RequirePackage[T1]{fontenc}
\documentclass[letterpaper, 10 pt, conference]{ieeeconf}  
\IEEEoverridecommandlockouts                              
\usepackage{mathptmx}
\DeclareMathAlphabet{\mathcal}{OMS}{cmsy}{m}{n}
\usepackage{cite}
\usepackage{amsmath,amssymb,amsfonts}
\allowdisplaybreaks
\usepackage{graphicx}
\usepackage{dsfont}
\usepackage{xcolor}
\usepackage{booktabs}
\usepackage{array}
\usepackage{tabularx}
\usepackage{microtype}
\usepackage{needspace}
\usepackage{url}
\usepackage{xurl}

\DeclareUrlCommand\email{\urlstyle{rm}}

\newcommand{\R}{\mathbf{R}}
\newcommand{\pos}{\mathbf{p}}
\newcommand{\T}{\mathbf{T}}
\newcommand{\X}{\mathbf{X}}
\newcommand{\Hm}{\mathbf{H}}
\newcommand{\Zm}{\mathbf{Z}}
\newcommand{\Bm}{\mathbf{B}}
\newcommand{\Mm}{\mathbf{M}}

\newcommand{\Log}{\operatorname{Log}}
\newcommand{\ind}{\mathbf{1}}

\newcommand{\cmark}{$\checkmark$}

\newcommand{\fvis}{$f_{\mathrm{vis}}$}
\newcommand{\fhand}{$f_{\mathrm{hand}}$}
\newcommand{\fpivot}{$f_{\mathrm{pivot}}$}
\newcommand{\fstatic}{$f_{\mathrm{stat}}$}

\newlength{\lhswd}
\newcommand{\lhs}[2][r^{\mathrm{grasp}}_{kl}]{%
  \settowidth{\lhswd}{$#1$}%
  \makebox[\lhswd][l]{$#2$}}

\begin{document}
\title{\LARGE \bf
VICON: Visual-Inertial-Contact based Hand-Object \\Tracking for Manipulation Datasets}

\author{Youbeen Jeon$^{1}$, Uiseong Shin$^{1}$, Hwanchul La$^{1}$, Jaeseong Kang$^{1}$, Hyelim Choi$^{2}$ and Yongseok Lee$^{\dagger}$%
\thanks{Youbeen Jeon, Uiseong Shin, Hwanchul La, Jaeseong Kang, and Yongseok Lee are with DGIST, Daegu 42988, Republic of Korea (e-mail:
        \protect\email{yoobin1liyfcb@gmail.com}; \protect\email{tlsdmltjd52@gmail.com};
        \protect\email{hc.la@dgist.ac.kr}; \protect\email{rkd020320@gmail.com};
        \protect\email{yslee@dgist.ac.kr}). Hyelim Choi is with Seoul National University, Republic of Korea (e-mail: \protect\email{helmchoi@snu.ac.kr}).}%
\thanks{$^{1}$These authors contributed equally to this work as first authors.}%
\thanks{$^{2}$Second author.}%
\thanks{$^{\dagger}$Corresponding author: Yongseok Lee (e-mail: \protect\email{yslee@dgist.ac.kr}).}%
}
\maketitle
\thispagestyle{empty}
\pagestyle{empty}
\bstctlcite{IEEEexample:BSTcontrol}

\begin{abstract}
Learning dexterous manipulation benefits from human demonstration datasets that capture diverse and natural hand-object interactions. In particular, contact points and forces provide supervision on where and how strongly to interact, which cannot be fully captured by motion trajectories alone. However, methods for jointly capturing hand and object motion, contact points, and forces remain limited. Moreover, severe occlusion from hand-object interaction challenges accurate tracking of both hands and objects. To address these limitations, we present a Visual-Inertial-CONtact based hand-object tracking (VICON) framework. It holistically captures both hand and object motion along with contact information during manipulation, even under severe occlusion. First, we adopt a visual-inertial glove~\cite{lee2021vist} and an RGB-D camera for accurate hand tracking, and redesign the glove to incorporate contact sensing. Specifically, force-sensitive resistors (FSRs) are placed on the glove based on human grasp frequency to synchronously record contact states and calibrated normal forces.
Second, without requiring pre-existing CAD models, we estimate object poses using RGB-D images and a mesh reconstructed from a monocular video. We propose factor-graph-based object trajectory estimation that fuses object-pose estimates weighted by visibility under hand-object occlusion, FSR measurements, and a hand-motion prior. Across 40 motion-capture sessions with five objects, VICON achieves a 2.5\% failed-frame rate compared with 50.9--64.6\% for the baselines, with median errors of 3.9\,mm and 3.0$^\circ$ under occlusion. Using VICON, we construct a dataset containing synchronized hand-object motion, contact points, and normal forces, and will publicly release an expanded dataset covering 10 object categories at \url{https://github.com/VICON-dataset/dataset}.

\end{abstract}

\section{Introduction}\label{sec:introduction}
Dexterous manipulation is a fundamental capability for humanoid robots operating in human environments, where everyday tasks require coordinated hand motion and continuous physical interaction with objects. Learning such intricate behaviors heavily relies on rich demonstrations.
However, collecting demonstrations via teleoperation is challenging; differences in kinematics and workspace between human and robot hands make real-time control unintuitive, severely limiting data collection efficiency and degrading motion quality. In contrast, directly capturing human demonstrations bypasses these difficulties, naturally preserving human dexterity and task variability across diverse objects and environments~\cite{hoque2026egodex,tao2025dexwild}. Yet, to fully realize this potential, hand-object interactions must be measured reliably.
Rich demonstration data requires more than hand-object motion alone, since successful dexterous manipulation fundamentally depends on where contact occurs and how much force is applied. Such information is particularly important for learning contact-rich manipulation policies~\cite{chen2025dexforce} and emerging force-aware vision-language-action models~\cite{yu2025forcevla}. Demonstration data should therefore capture contact points and forces alongside hand and object motion. However, many existing datasets provide detailed hand-object geometry and contact annotations without directly measuring the applied forces~\cite{brahmbhatt2020contactpose,fan2023arctic}. Although recent tactile datasets and sensing systems provide richer contact information~\cite{sundaram2019stag,song2025opentouch,ni2026tactidex}, simultaneously capturing hand-object poses along with contact locations and forces remains challenging within a compact setup. In particular, achieving accurate ground truth often requires multi-view camera systems, restricting data collection to dedicated environments.

\begin{figure}[t]
\centering
\includegraphics[width=\columnwidth]{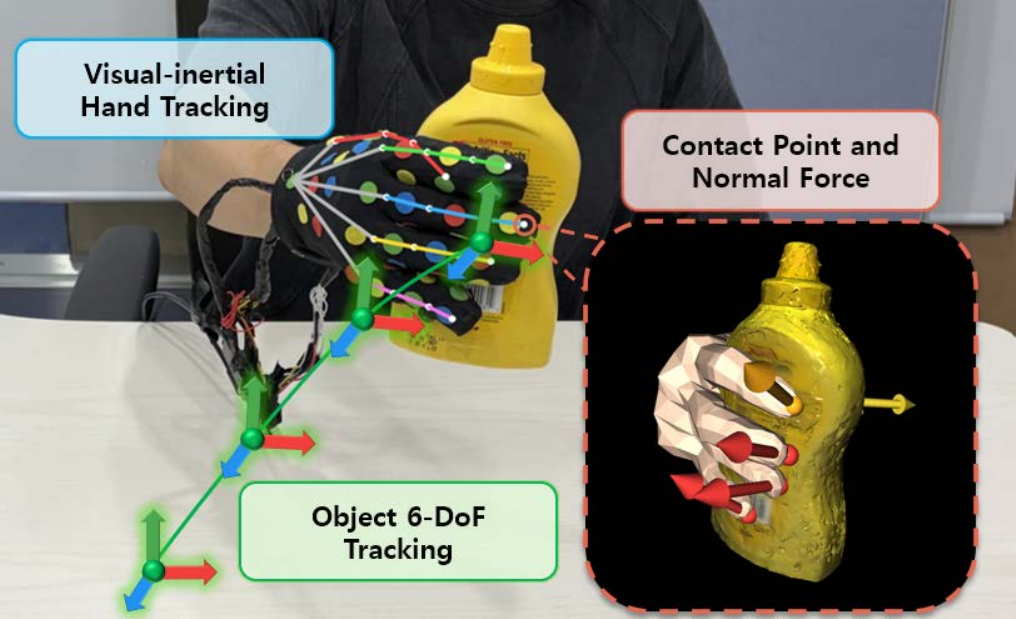}
\caption{VICON records hand motion, object pose, contact points, and forces using a visual-inertial glove~\cite{lee2021vist} with integrated FSRs and a single RGB-D camera. The glove tracks the hand skeleton, and our offline refinement maintains robust and accurate object 6-DoF tracking even under severe occlusion. Inset: object-frame contact points, with arrows colored by calibrated normal force magnitude.}
\label{fig:teaser}
\end{figure}

With a compact setup, occlusion presents another major challenge for reliable hand-object tracking. Under severe occlusion, even state-of-the-art vision-based pose estimators can become unreliable or fail entirely. Although existing works have attempted to mitigate occlusion in hand pose estimation~\cite{park2022handoccnet} and hand-object tracking~\cite{sridhar2016joint,qi2024hoisdf,fan2024hold}, these approaches still rely heavily on visual cues, which can become unreliable under severe or prolonged occlusion. Recent work has explored using hand motion as an additional constraint for object tracking~\cite{shin2026compose}, but maintaining robust tracking when visual evidence is largely unavailable remains challenging.

Wearable hand tracking directly captures hand kinematics, providing reliable motion estimates where vision-only estimation fails. In particular, visual-inertial skeletal tracking (VIST)~\cite{lee2021vist} combines inertial measurements with visual marker observations to robustly track hand motion under severe occlusions. Because hand and object trajectories are tightly coupled during manipulation, this robust hand motion can serve as a prior to guide object pose estimation when visual object cues become unreliable. However, this kinematic constraint is physically valid only when the hand and object are in direct contact. Contact sensing therefore serves as a crucial gate to determine when the hand-motion prior should be applied. To address this, we integrate force-sensitive resistors (FSRs) directly onto the glove, providing occlusion-invariant contact detection while simultaneously yielding calibrated normal force measurements.

Building on these components, we present \textbf{VICON} (Fig.~\ref{fig:teaser}), a multimodal tracking framework built on the VIST glove~\cite{lee2021vist} integrated with eight FSRs, together with a single RGB-D camera. An offline factor graph fuses object-pose estimates from FoundationPose~\cite{wen2024foundationpose}, weighted by object visibility, with a contact-gated hand-motion prior to refine object trajectories under occlusion. Using the refined trajectories, contact points are located on the object surface, which are paired with calibrated FSR forces. The main contributions of our work are:

\begin{enumerate}
   \item \textbf{Visual-inertial hand tracking system with contact sensing.}

        We integrate eight FSRs into the VIST glove~\cite{lee2021vist}, positioned according to human grasp contact frequency~\cite{gonzalez2014hand}. Together with a single RGB-D camera, the resulting \textbf{VICON-glove} synchronously records hand motion, contact states, and calibrated normal contact forces.

\item \textbf{Robust hand-object tracking algorithm under severe occlusion.}

   We fuse a contact-gated hand-motion prior with visibility-weighted object pose estimates to refine object trajectories and maintain robust tracking even under severe occlusion.

\item \textbf{Scalable contact-rich dataset collection for custom objects.}

We present a pipeline that reconstructs object geometry from monocular video and captures synchronized hand motion, object pose, object-frame contact points, and calibrated normal forces using only a VICON-glove and a single RGB-D camera. We will publicly release an expanded dataset covering 10 object categories.
\end{enumerate}

\section{Related Works}\label{sec:related}
\subsection{Vision-Based Hand and Object Pose Estimation}
Vision-based methods have been widely studied for recovering hand and object poses during manipulation from RGB or RGB-D observations.
For object pose estimation, MegaPose~\cite{labbe2022megapose} estimates the 6-DoF pose of novel objects using rendered CAD views for coarse pose selection and iterative refinement.
FoundationPose~\cite{wen2024foundationpose} supports both estimation and tracking with CAD models or neural object representations learned from reference RGB-D images, although occlusion degrades the visual cues needed for reliable alignment.
A second line of work jointly estimates hand and object poses by aligning hand-object models with RGB-D observations using proximity-based contact and occlusion constraints~\cite{sridhar2016joint}. HOLD~\cite{fan2024hold} reconstructs both from monocular video using neural implicit surfaces and contact constraints without an object template.
More recently, ComPose~\cite{shin2026compose} learns from monocular RGB video to weight hand joints and combine hand- and object-derived motion cues for object tracking.
Despite these advances, image-derived geometry and motion consistency may not reliably distinguish near-contact state from actual contact at an occluded surface, and none of these methods directly measures contact forces.

\begin{table}[t]
\centering
\caption{Comparison of our dataset with \\existing interaction datasets}
\label{tab:datasets}
\footnotesize
\setlength{\tabcolsep}{1.6pt}
\renewcommand{\arraystretch}{1.1}
\begin{tabularx}{\columnwidth}{@{}lcc>{\hsize=.85\hsize\centering\arraybackslash}X>{\hsize=.8\hsize\centering\arraybackslash}X>{\hsize=1.35\hsize\centering\arraybackslash}X@{}}
\toprule
Dataset & Hand & Object & Contact & Force/ & Setup \\
 & pose & pose & format & tactile & \\
\midrule
EgoDex~\cite{hoque2026egodex} & \cmark & -- & -- & -- & Vision Pro \\
ContactPose~\cite{brahmbhatt2020contactpose} & \cmark & \cmark & Object surface map & -- & 3 RGB-D, thermal, mocap \\
ARCTIC~\cite{fan2023arctic} & \cmark & \cmark & Object surface map & -- & 8+1 RGB, mocap \\
OpenTouch~\cite{song2025opentouch} & \cmark & -- & Hand map & Pressure & Aria glasses, tracking and tactile gloves \\
TactiDex~\cite{ni2026tactidex} & \cmark & \cmark & Hand map & Pressure & Mocap, tracking and tactile gloves \\
ForceVLA-Data~\cite{yu2025forcevla} & -- & -- & -- & TCP wrench (est.) & Robot arm, 2 RGB-D \\
\midrule
VICON (ours) & \cmark & \cmark & Object-frame 3D points & Normal force (N) & 1 RGB-D, 1 VICON-glove (8 FSRs) \\
\bottomrule
\end{tabularx}
\par\vspace{4pt}
\begin{minipage}{\columnwidth}\scriptsize\raggedright
\cmark: provided; --: not provided. Contact format: stored representation of contact (ContactPose: thermal imprint after a static grasp; ARCTIC: derived from geometry). Force/tactile: recorded signal type.
\end{minipage}
\vspace{-4pt}
\end{table}

\subsection{Hand-Tracking Gloves and Contact Sensing}
Sensorized gloves have been developed to capture hand motion and tactile interactions during manipulation.
VIST~\cite{lee2021vist} fuses glove-mounted IMUs with visual markers under hand kinematic constraints, using inertial propagation and visual corrections to maintain hand tracking through occlusion.
A markerless approach uses inertial pose priors to infer image keypoints and their uncertainty and combines the learned keypoint likelihoods with inertial observations in a factor graph~\cite{choi2026robusthand}.
FSGlove~\cite{li2025fsglove} uses IMUs placed on the glove, whose alignment is calibrated from reference poses and fits a personalized hand shape using predefined finger contact constraints. However, an external dorsal tracker is additionally used to recover the hand's global pose.
For contact sensing, STAG~\cite{sundaram2019stag} records spatial patterns of normal force using 548 piezoresistive sensors distributed over the hand, and learns to identify objects and estimate their weights from these tactile signatures.
However, neither kinematic tracking nor tactile sensing is sufficient on its own; kinematic tracking cannot measure applied force, while tactile sensing requires hand-object kinematics to locate contacts on the object surface. Despite this clear complementarity, prior works rarely integrate both modalities in a single wearable system.

\subsection{Hand-Object Interaction Datasets for Manipulation}
Table~\ref{tab:datasets} compares existing datasets in terms of pose annotations, contact and force signals, and recording setups.
EgoDex~\cite{hoque2026egodex} pairs large-scale egocentric video with hand poses but has no object pose, contact, or force.
ContactPose~\cite{brahmbhatt2020contactpose} maps thermal imprints after static grasps onto object meshes alongside hand and object poses.
ARCTIC~\cite{fan2023arctic} derives contact from reconstructed hand-object geometry during bimanual manipulation of articulated objects, requiring a costly multi-camera motion-capture studio.

Tactile datasets pair measured pressure with hand motion.
OpenTouch~\cite{song2025opentouch} combines full-hand pressure, hand tracking, and egocentric video, but provides no object poses.
More recently, TactiDex~\cite{ni2026tactidex} combines dense tactile gloves with hand-object motion capture, using pressure-gated proximity and tactile-constrained optimization, but requires a motion-capture workspace.

Force-aware robot datasets record external loads for contact-rich policy learning.
ForceVLA~\cite{yu2025forcevla} synchronizes images, proprioception, and estimated six-axis tool-center-point (TCP) wrenches in ForceVLA-Data across five contact-rich tasks and fuses them in a vision-language-action policy. However, these wrenches describe net end-effector loads without resolving individual human hand-object contacts.

In contrast to these systems, VICON enables scalable collection of contact-rich demonstrations. Specifically, it is low-cost and wearable, since the VICON-glove builds on the VIST glove, whose material cost is about \$100 and weight is 52--55\,g~\cite{lee2021vist}, and adds only eight FSRs. Moreover, it requires only one RGB-D camera and no motion-capture studio, so recording is not restricted to a dedicated capture space. In addition, it provides accurate data even under severe occlusion, as the glove tracks the hand robustly and the FSRs measure contact directly (Sec.~\ref{sec:baselines}). Finally, it pairs object-frame contact points with calibrated normal forces, a feature absent from prior datasets in Table~\ref{tab:datasets}. As the volume of demonstration data becomes increasingly important for learning manipulation, this combination of low acquisition cost and high accuracy is a key advantage over costly studio-based systems such as ARCTIC~\cite{fan2023arctic}.

\section{Hardware System}\label{sec:hardware}
The hardware consists of the VICON-glove (Fig.~\ref{fig:sensor-placement}) and one Intel RealSense D435i camera. The glove is built on VIST~\cite{lee2021vist}, which incorporates IMUs and visual markers, and is augmented with eight FSRs. The camera is positioned beside the subject at roughly eye level, providing a side view of the workspace; depth is computed from the infrared pair with Fast-FoundationStereo~\cite{wen2026fastfoundationstereo} at about 30\,Hz.

\subsection{Hand-Tracking Glove}
The glove consists of two layers, an inner glove carrying the IMUs and an outer glove carrying the markers. Fourteen IMUs are mounted on the inner glove: one on the dorsum of the hand and thirteen along the fingers, with no IMUs on the distal phalanges of the ring and little fingers (Fig.~\ref{fig:sensor-placement}(a)). The outer glove holds 80 markers in four colors (Fig.~\ref{fig:sensor-placement}(b)), and visual observations of these markers are used to correct drift in the inertial estimates.

\subsection{FSR Placement}
\begin{figure}[t]
\centering
\begin{minipage}[t]{0.56\columnwidth}
\centering
\includegraphics[width=\linewidth]{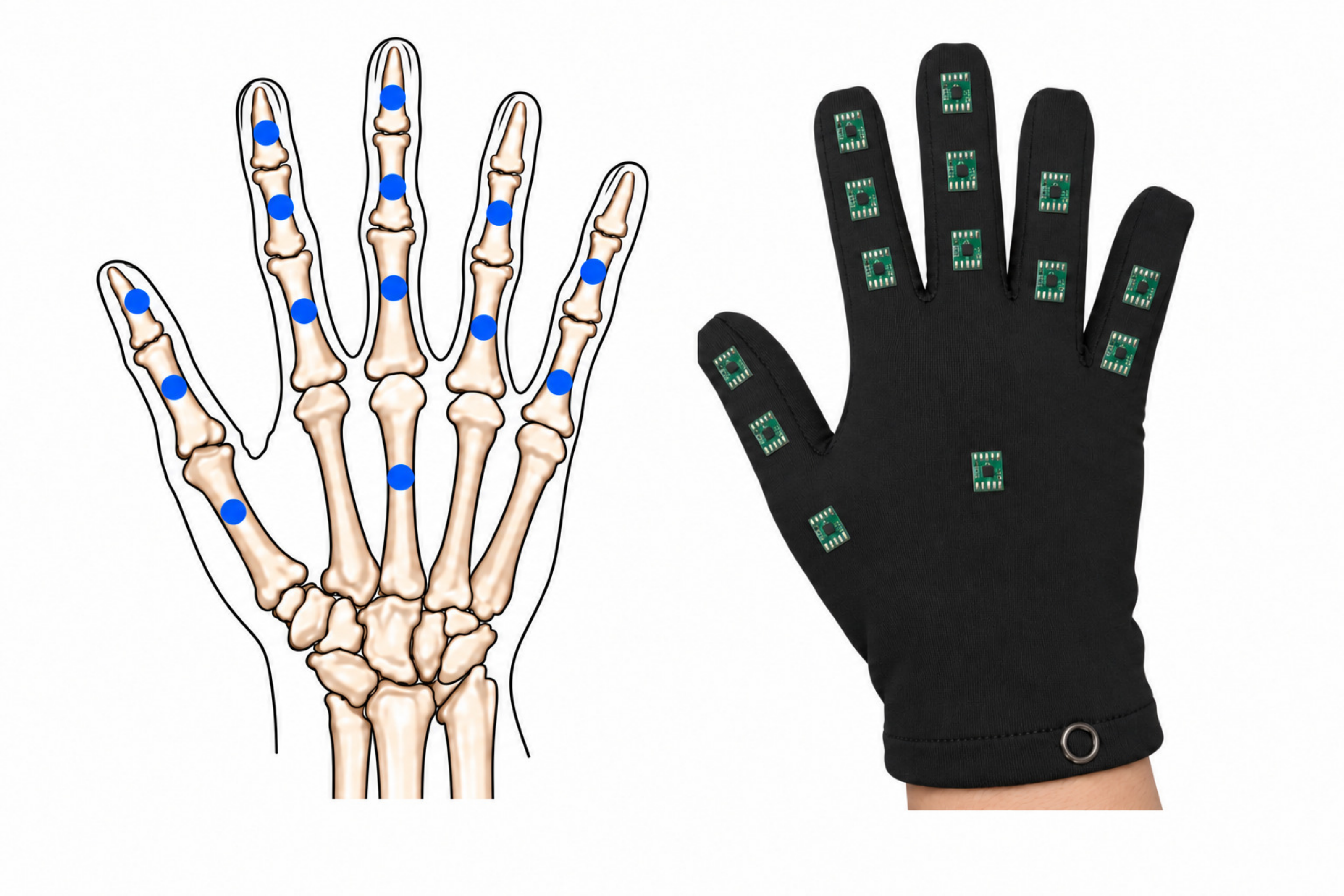}
\par\smallskip
{\small (a)}
\end{minipage}
\hspace{2mm}
\begin{minipage}[t]{0.27\columnwidth}
\centering
\includegraphics[width=\linewidth]{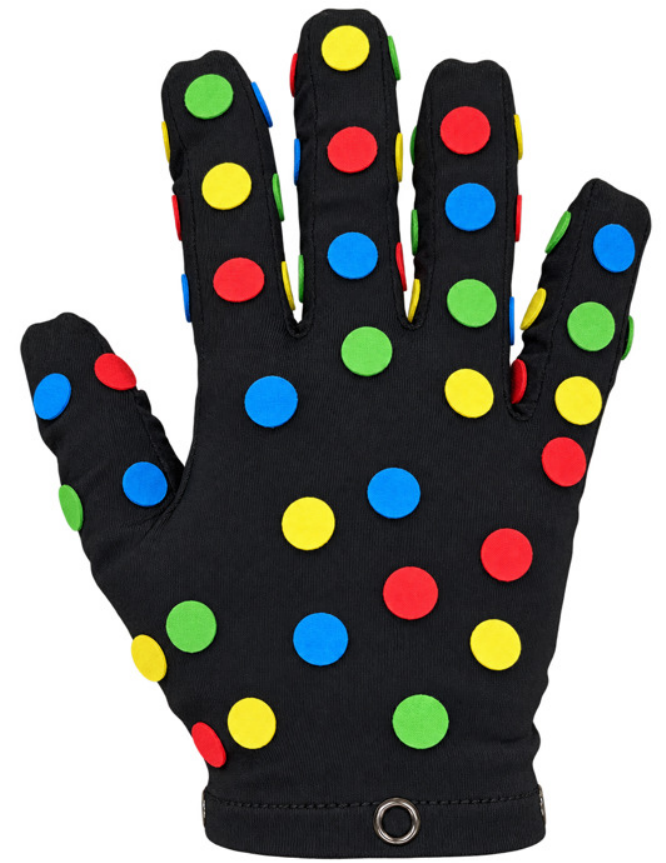}
\par\smallskip
{\small (b)}
\end{minipage}

\vspace{1.5mm}

\begin{minipage}[t]{0.5\columnwidth}
\centering
\includegraphics[width=\linewidth]{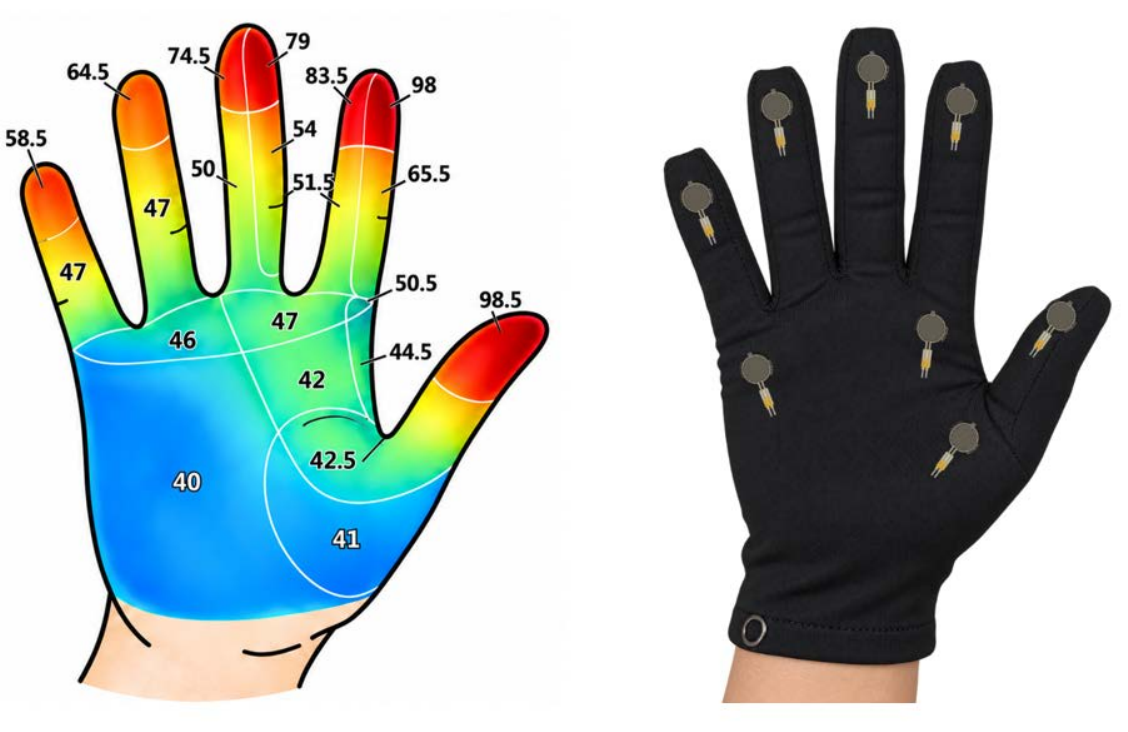}
\par\smallskip
{\small (c)}
\end{minipage}
\hspace{2mm}
\begin{minipage}[t]{0.27\columnwidth}
\centering
\includegraphics[width=\linewidth]{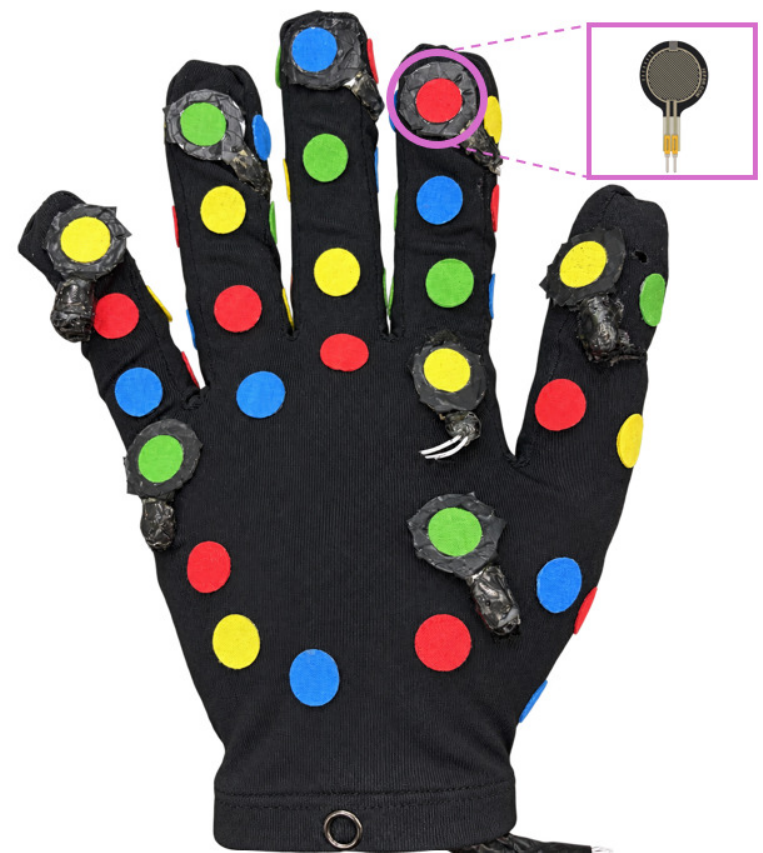}
\par\smallskip
{\small (d)}
\end{minipage}

\caption{Multi-layer VICON-glove. (a)~Placement of the 14 IMUs in the inner glove. (b)~Outer glove with the visual markers. (c)~Hand contact-frequency map~\cite{gonzalez2014hand}, averaged over the mechanic and housemaid tasks (values in \%), used to choose the FSR sites. (d)~Palmar view of the eight FSRs.}
\label{fig:sensor-placement}
\end{figure}

We determine the FSR placement based on a contact-frequency analysis of human grasps(Fig.~\ref{fig:sensor-placement}(c))~\cite{gonzalez2014hand}. Their study classifies grasps into power, precision, and intermediate grasps and analyzes contact frequencies during mechanic and housemaid tasks. The results show that the fingertips exhibit the highest contact frequencies, while frequent contact is also observed at the thenar eminence and distal palm. Based on these observations, we place a sensor on each of the five fingertips, one on the thenar eminence, and two on the distal palm over the second- and fifth-metacarpal regions (Fig.~\ref{fig:sensor-placement}(d)). For the ulnar-side distal-palm sensor, the analysis reports contact frequency as a single region spanning the third to fifth metacarpals, without distinguishing the areas beneath the ring and little fingers~\cite{gonzalez2014hand}. We therefore place the FSR near the little-finger metacarpophalangeal(MCP) joint on the palmar side to provide broader spatial coverage within this region.

\begin{figure*}[t]
\centering
\includegraphics[width=0.9\textwidth]{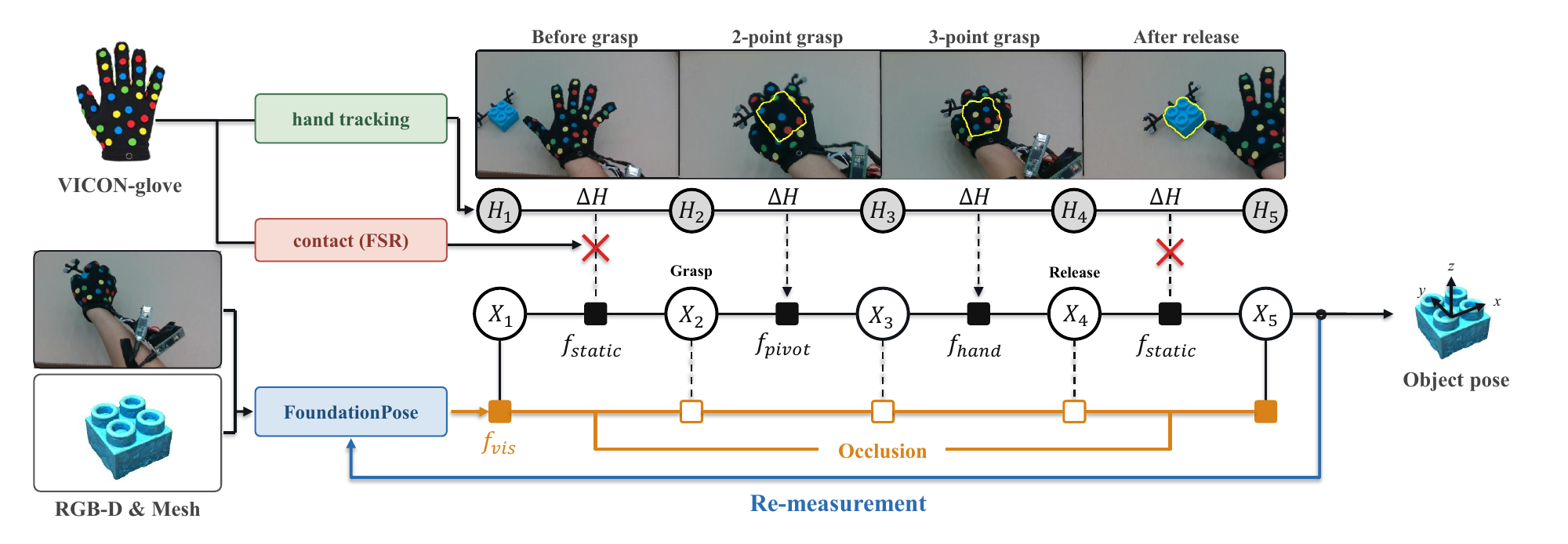}
\caption{VICON on a recorded session. The VICON-glove provides hand tracking (poses $\Hm_1$--$\Hm_5$, increments $\Delta H$) and contact (FSR); the RGB-D camera feeds FoundationPose, whose candidates enter as \fvis\ (filled: accepted; hollow: rejected under occlusion). Between object poses $X_k$, contact selects \fstatic\ (visible, after release; red crosses: rejected hand increments), \fpivot\ (2-point grasp), or \fhand\ (3-point grasp). Re-measurement feeds the refined poses back to FoundationPose.}
\label{fig:method-paths}
\end{figure*}

\section{Visibility- and Contact-Guided \\Object Pose Refinement}\label{sec:method}

\subsection{Notation and Overview}\label{sec:recording}

Throughout, $\T_{AB}\in SE(3)$ denotes the pose of frame $B$ expressed in frame $A$. We use the stationary camera frame $C$, the hand frame $H$ (the palm frame), and the object frame $O$ attached to the mesh. We estimate the object poses $\X_k=\T_{CO,k}=(\R_k,\pos_k)$ over the frames $k$ of a session, while the hand poses $\Hm_k=\T_{CH,k}$ and the centers $\mathbf s^{C}_{i,k}$ of the eight FSRs ($i=1,\dots,8$), where a superscript denotes the frame in which a point is expressed, are provided by the glove and treated as fixed inputs during optimization.

Offline refinement takes as input the initial FoundationPose trajectory $\{\widetilde\X_k\}$, the recorded RGB-D frames, the hand trajectory, and the FSR streams (Fig.~\ref{fig:method-paths}). FoundationPose is initialized from a \emph{seed pose} ($\widetilde\X_k$ at first, the refined trajectory later) and returns a \emph{candidate} $\Zm_k$. Candidates are evaluated by visibility (Sec.~\ref{sec:visibility}). Contact extraction then alternates with refinement (Sec.~\ref{sec:contact}): each round begins with a \emph{re-measurement}, re-running FoundationPose bi-directionally from the seed poses (pose-refinement network only), and ends in a factor graph fusing the visibility-weighted candidates with \fhand, \fpivot, and \fstatic\ (Sec.~\ref{sec:batch}). Contact points and forces follow from the refined poses (Sec.~\ref{sec:proximity}).

\subsection{Visual Evidence and Visibility Weighting}\label{sec:visibility}

Since FoundationPose keeps outputting poses under occlusion even when little of the object is visible, we score each candidate pose $\Zm_k$ by comparing the rendered mesh depth $D_O$ with the observed depth $D$. Let $\Omega$ denote the set of pixels with $D_O(u),D(u)>0.001$, and let $\delta(u)=D(u)-D_O(u)$, with depths in meters. We define
\begin{align}
v_d(\Zm_k)&=\frac{1}{|\Omega|}\sum_{u\in\Omega}\ind[\,|\delta(u)|<0.010\,],\label{eq:vis}\\
v_{\mathrm{far}}(\Zm_k)&=\frac{1}{|\Omega|}\sum_{u\in\Omega}\ind[\,\delta(u)>0.020\,],\label{eq:far}
\end{align}
where $\ind[\cdot]$ is the indicator function. The visibility $v_d$ falls under either occlusion or pose error, reducing the measurement weight, while $v_{\mathrm{far}}$ flags candidates floating in front of the measured surface.

Before re-measurement, the tracked hand skeleton and forearm are projected as capsules to form a hand mask; depths inside it more than 20\,mm in front of the seed rendering are removed, with a dilated margin, from the depth image passed to FoundationPose; the returned candidates are scored against the original depth.

A candidate is accepted only if $|\Omega|\geq200$, $v_d\geq0.10$, it does not penetrate the table by more than 25\,mm, and $v_{\mathrm{far}}$ is below a threshold that defaults to 0.15 and is adapted per session from high-visibility candidates. A candidate must then persist over consecutive frames without jumping from the previous accepted pose, agree with the anchors of its contact-free interval, and, during grasp, stay near the propagated pose $\Bm_k$.

For accepted measurements, the deviations $\sigma_p$ and $\sigma_r$ (Table~\ref{tab:sigma}) are interpolated between visibility knots with a cubic smoothstep and adjusted per candidate: $\sigma_p$ grows with displacement from the seed pose, penetration of the table, and, during a grasp, the distance the hand travels in the tracker's $100$\,ms lag, while heavy hand coverage raises the floor of both deviations. Below a visibility threshold (0.50 by default, lower for temporally stable candidates), only the depth-supported tilt is retained with its own deviation and the twist about that axis keeps a weak but finite weight. These gates, masks, and covariances define the \emph{visibility weighting} in \fvis.

\begin{table}[t]
\caption{Nominal visual-measurement uncertainty\\as a function of visibility}
\label{tab:sigma}
\centering
\footnotesize
\setlength{\tabcolsep}{1.4pt}
\renewcommand{\arraystretch}{1.1}
\begin{tabular*}{\columnwidth}{@{\extracolsep{\fill}}lccccccc@{}}
\toprule
$v_d$ & 0.10 & 0.15 & 0.30 & 0.50 & 0.70 & 0.90 & 1.0 \\
\midrule
$\sigma_p$ (mm) & 6 & 5 & 3 & 3 & 2 & 1.2 & 0.8 \\
$\sigma_r$ (deg) & 180 & 180 & 8 & 3 & 2 & 1 & 0.8 \\
\bottomrule
\end{tabular*}
\par\vspace{4pt}
\begin{minipage}{\columnwidth}\fontsize{8}{9}\selectfont\raggedright
Nominal standard deviations of the visual measurements before the per-candidate adjustments of Sec.~\ref{sec:visibility}; $180^\circ$ denotes a weak but finite rotational weight.
\end{minipage}
\vspace*{-4pt}
\end{table}

\subsection{Tactile Contact Timing and Hand-Motion Prior}\label{sec:contact}

FSR activation and FSR--object proximity determine whether the \emph{hand-motion prior} is applied at each frame. Each FSR signal is sampled at 200\,Hz, low-pass filtered at 20\,Hz, and converted to a per-frame binary activation $a_{i,k}\in\{0,1\}$ using a Schmitt trigger with a short hold time. Let $d_{i,k}$ be the unsigned distance from the center of FSR $i$ to the mesh at a reference pose, which is kept fixed while the contact states are extracted. FSR $i$ is \emph{recognized} at frame $k$ when $a_{i,k}=1$ and $d_{i,k}\le0.030$; writing $\rho_{i,k}=a_{i,k}\,\ind[d_{i,k}\leq0.030]$, we declare a grasp when at least three FSRs are recognized,
\begin{equation}
c_k=\ind\!\left[\sum_{i=1}^{8}\rho_{i,k}\;\geq3\right].
\label{eq:contact}
\end{equation}

Because the reference poses carry their own error, contact states are extracted twice. The first pass registers a contact when an FSR activates within 100\,mm of the initial poses and keeps it until it deactivates, so that intervals survive frames with inaccurate initial poses; a first refinement then yields a trajectory accurate enough to serve as reference. In the second pass, contact states are re-extracted from that trajectory using~\eqref{eq:contact}, and a second refinement yields all results below. We call the criterion~\eqref{eq:contact} \emph{tactile contact timing} (contact timing for short); it only gates when the prior applies and does not model force magnitude or force closure. Frames with $c_k=1$ define grasp intervals, in which the object moves rigidly with the hand (\fhand). Adjacent runs of at least five frames with exactly two recognized fingertip contacts (2-point grasp), which occur during grasp formation, release, or weakened contact, are also propagated with the hand, with the prior relaxed about the line through the two contacts (\fpivot).

Alongside this contact gate, we maintain a propagated pose $\Bm_k$. We call accepted measurements with $v_d\geq0.7$ \emph{anchors}; $\Bm_k$ is initialized from the first anchor and, while the object is not grasped, updated at every anchor. During a grasp with valid hand tracking (a \emph{usable hand increment}), it is propagated with the hand, $\Bm_k=\Hm_k\Hm_{k-1}^{-1}\Bm_{k-1}$, and remains fixed when neither a visual update nor a usable hand increment is available. Across \emph{visual gaps}, runs of frames without accepted measurements, the propagated pose is reconciled with the anchors on either side: agreeing anchors indicate a stationary object, while a mismatch is distributed over the hand-propagated frames. A grasp interval without any anchor is assigned synthetic poses $\bar\Zm_k$ from the propagated trajectory, with large nominal deviations of 80\,mm and 45$^\circ$.

\subsection{Object-Pose Factor Graph}\label{sec:batch}

These components are combined in a single factor graph over $\{\X_k\}$, while the hand poses, contact states, and object geometry are kept fixed. Let $\Mm_k=\Bm_{k-1}^{-1}\Bm_k$ denote the increment of the propagated pose from frame $k-1$ to $k$, expressed in the object frame. It equals the hand motion, plus the anchor correction where it spans a visual gap. The residuals of the visual, synthetic, hand-motion, stationary, and grasp-link factors are defined as
\begin{align}
\lhs{r^{\mathrm{vis}}_k}&=\Log(\Zm_k^{-1}\X_k),\\
\lhs{r^{\mathrm{syn}}_k}&=\Log(\bar\Zm_k^{-1}\X_k),\label{eq:unary}\\
\lhs{r^{\mathrm{hand}}_k}&=\Log(\Mm_k^{-1}\X_{k-1}^{-1}\X_k),\label{eq:hand}\\
\lhs{r^{\mathrm{stat}}_k}&=\Log(\X_{k-1}^{-1}\X_k),\label{eq:rest}\\
\lhs{r^{\mathrm{grasp}}_{kl}}&=\Log(\X_k^{-1}\Hm_k\Hm_l^{-1}\X_l),\label{eq:grasp}
\end{align}
where $\Log:SE(3)\to\mathbb R^6$ uses rotation-first tangent coordinates, and the grasp-link residual enforces a constant hand-to-object transform $\Hm_k^{-1}\X_k$ across linked frames. With $\|r\|_A=(r^TAr)^{1/2}$ and the Huber loss $\phi$ (threshold 1.345), we minimize
\begin{align}
&\sum_{k\in\mathcal V}\phi\big(\|r^{\mathrm{vis}}_k\|_{\Sigma_k^{-1}}\big)
+\sum_{k\in\bar{\mathcal V}}\phi\big(\|r^{\mathrm{syn}}_k\|_{\Lambda_k^{-1}}\big)\nonumber\\
&+\tfrac12\sum_{k\in\mathcal H}\|r^{\mathrm{hand}}_k\|_{Q_k^{-1}}^2
+\tfrac12\sum_{k\in\mathcal W}\|r^{\mathrm{stat}}_k\|_{W_k^{-1}}^2\nonumber\\
&+\tfrac12\sum_{(k,l)\in\mathcal A}\|r^{\mathrm{grasp}}_{kl}\|_{P_{kl}^{-1}}^2\nonumber\\
&+\sum_{(k,l)\in\mathcal G}\phi\big(\|r^{\mathrm{grasp}}_{kl}\|_{U_{kl}^{-1}}\big)
\label{eq:objective}
\end{align}
The sets $\mathcal V$, $\bar{\mathcal V}$, $\mathcal H$, $\mathcal W$, $\mathcal A$, and $\mathcal G$ index accepted measurements, synthetic poses, usable hand increments, stationary pairs $(k-1,k)$, adjacent grasp pairs without a usable hand increment, and the long-range grasp links described below.

Except for the pivot relaxation below, all $6\times6$ covariances are block-diagonal in rotation and translation. The visual covariance $\Sigma_k$ uses the adjusted per-candidate deviations of Sec.~\ref{sec:visibility}, and the synthetic-pose covariance $\Lambda_k$ is isotropic within each block. The base hand-motion covariance $Q_k^{0}$, stationary covariance $W_k$, and adjacent grasp-link covariance $P_{kl}$ are diagonal, with translation/rotation deviations of 0.55\,mm/0.2$^\circ$, 0.01\,mm/0.03$^\circ$, and 1.5\,mm/1.5$^\circ$ per nominal frame interval $\Delta t_{\mathrm{nom}}$, rescaled by $\Delta t_k/\Delta t_{\mathrm{nom}}$, visual disagreement, occlusion, and hand-tracking quality. For grasps of at least ten frames, long-range links connect each grasp frame to the anchor frames on either side, with an 8$^\circ$ rotational deviation in $U_{kl}$ and translation left effectively unconstrained. For each pair in a 2-point grasp, let $\mathbf s^{O}_{j,k-1}=\Bm_{k-1}^{-1}\mathbf s^{C}_{j,k-1}$, $j\in\{a,b\}$, denote the two recognized FSR centers in the object frame, $\mathbf n_{k-1}$ the unit vector along the line connecting them, and $\boldsymbol\mu_{k-1}$ its midpoint. The rigid hand-motion prior is relaxed about this line through a rank-one covariance update,
\begin{equation}
Q_k^{\mathrm{piv}}=Q_k^{0}+(\sigma_{\mathrm{free}}\alpha_k)^2\,\mathbf u_k\mathbf u_k^{T},
\label{eq:pivot-cov}
\end{equation}
with $\mathbf u_k=[\,\mathbf n_{k-1};\,\boldsymbol\mu_{k-1}\times\mathbf n_{k-1}\,]$ the screw axis of rotation about the contact line, expressed in the object frame as in \eqref{eq:hand}, $\alpha_k=\max(\Delta t_k/\Delta t_{\mathrm{nom}},\,0.2)$, and $\sigma_{\mathrm{free}}=15^\circ$ per nominal frame interval. It applies only to pairs with a valid axis and $\Delta t_k\leq0.25$\,s, lowering the penalty on rotation about the contact line and its coupled translation. In \eqref{eq:objective}, $Q_k=Q_k^{\mathrm{piv}}$ for these pairs (\fpivot) and $Q_k=Q_k^{0}$ otherwise (\fhand).

The graph is initialized with the propagated poses and is solved in two stages using Levenberg--Marquardt in GTSAM. The first stage uses a synthetic-pose rotational deviation of $5^\circ$, and the second stage uses the nominal value of $45^\circ$.

\subsection{Contact and Force Annotation}\label{sec:proximity}

The 12-bit ADC readings from the FSRs are first converted into a dimensionless response and then mapped to normal force by a per-FSR calibration,
\begin{equation}
x = \frac{\mathrm{ADC}}{\max(4095-\mathrm{ADC},\,1)},
\qquad
F_i = f_i(x).
\label{eq:force-calibration}
\end{equation}
Because the datasheet specifies $\pm6\%$ part-to-part repeatability~\cite{interlink_fsr400}, each FSR was calibrated separately against a digital scale over slow press--release cycles up to $61\,\mathrm{N}$. Each $f_i$ is an affine fit $a_ix+b_i$ or a power law $a_ix^{n_i}$, whichever gives the higher $R^2$, joined linearly to the origin below the lowest calibrated point. The fits yield $R^2=0.83$--$0.99$ and mean absolute errors of $0.74$--$2.87\,\mathrm{N}$ overall and $0.68$--$2.50\,\mathrm{N}$ below $20\,\mathrm{N}$, where most recorded forces lie (Sec.~\ref{sec:dataset}). No FSR saturated within the calibrated range.

For contact annotation, each recognized FSR is associated with the nearest point on the object mesh $\mathcal{M}_O$ at the refined pose:
\begin{equation}
\mathbf{q}^{O}_{i,k}
\in
\arg\min_{\mathbf{q}\in\mathcal{M}_O}
\left\|
\X_k \mathbf{q} - \mathbf{s}^{C}_{i,k}
\right\|_2,
\qquad
\mathbf{q}^{C}_{i,k} = \X_k \mathbf{q}^{O}_{i,k}.
\label{eq:contact-point}
\end{equation}
Each record stores $\mathbf q^{O}_{i,k}$ and $\mathbf q^{C}_{i,k}$, the FSR--surface distance signed negative inside the mesh, and the synchronized force $F_{i,k}=f_i(x_{i,k})$.

\section{Experiments and Results}\label{sec:experiments}
\subsection{Recording Setup and Object Models}\label{sec:setup}
\begin{figure}[t]
    \centering
    \begin{minipage}[t]{0.45\columnwidth}
        \centering
        \includegraphics[width=\linewidth]{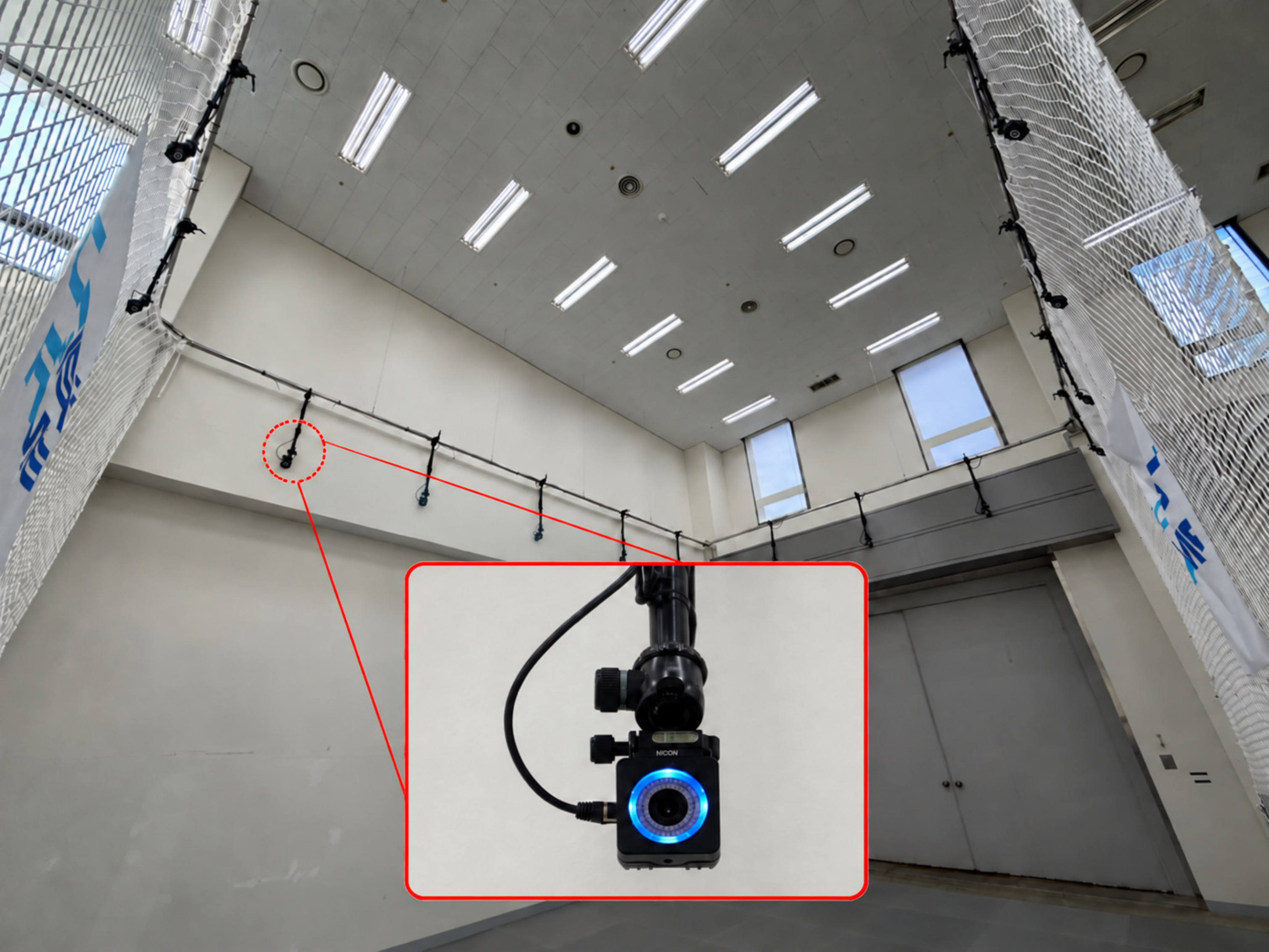}
        \par\smallskip
        {\footnotesize (a)}
    \end{minipage}
    \hfill
    \begin{minipage}[t]{0.45\columnwidth}
        \centering
        \includegraphics[width=\linewidth]{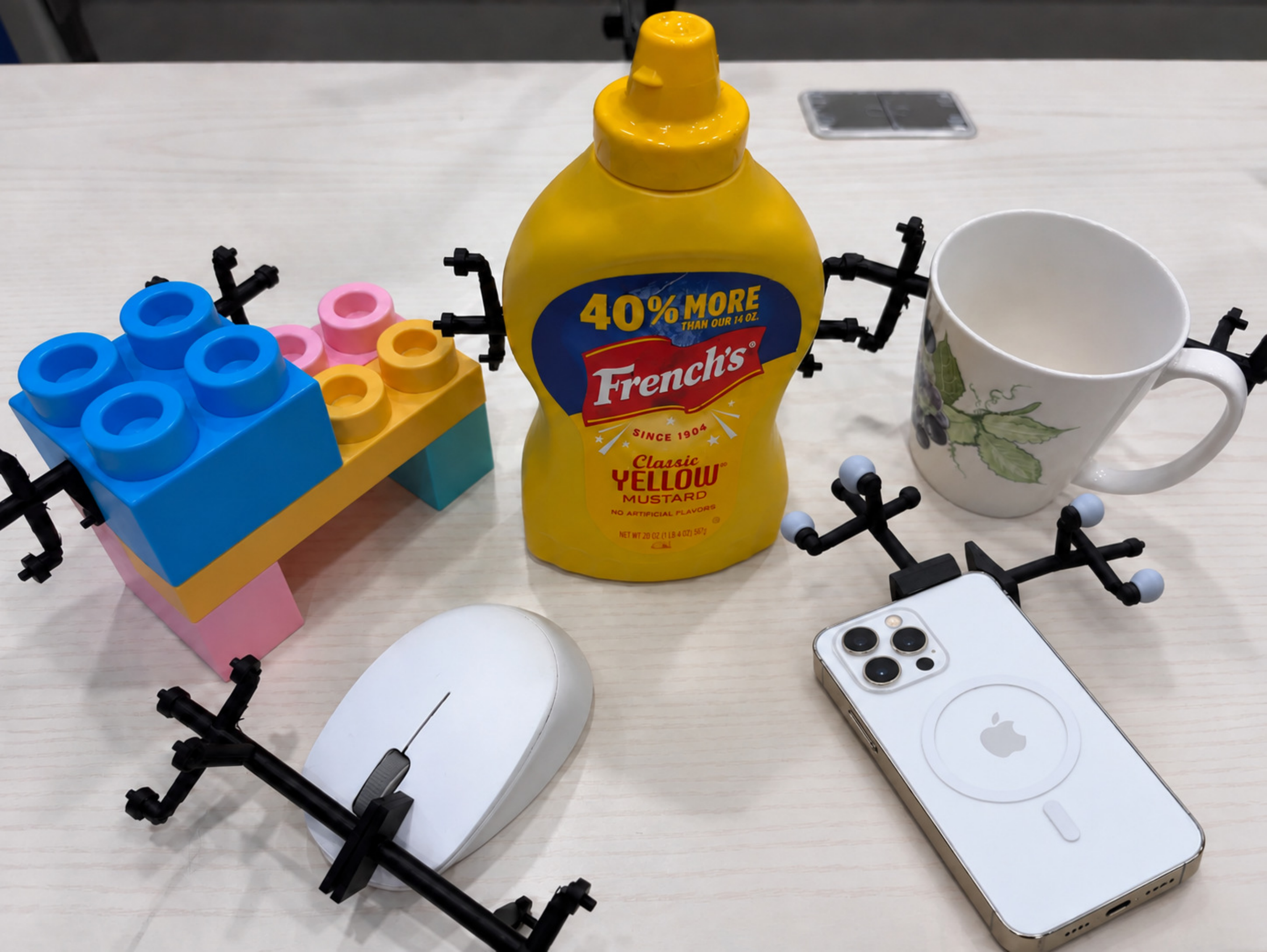}
        \par\smallskip
        {\footnotesize (b)}
    \end{minipage}
    \caption{Experimental setup. (a)~Motion-capture room; the inset shows one of the motion-capture cameras. (b)~The five objects: block, mouse, mug, mustard bottle, and phone.}
    \label{fig:contact-force-evaluation}
\end{figure}

Four subjects, each wearing a single glove, recorded 71 sessions with five rigid objects: a block, a mouse, a mug, a mustard bottle, and a phone (Fig.~\ref{fig:contact-force-evaluation}(b)), and each session consisted of approach, grasp, slow transport, and placement. Forty sessions (i.e., eight per object) were captured in a motion-capture room for ground truth (Fig.~\ref{fig:contact-force-evaluation}(a)); the other 31 were recorded without motion capture. The object meshes were obtained from monocular video via COLMAP~\cite{schonberger2016sfm}, 3DGS~\cite{kerbl2023gs}, and SuGaR~\cite{guedon2024sugar}, with a ChArUco board setting the metric scale and unobserved surfaces cleaned away; all evaluated object pose trackers use the same meshes. Validating on YCB objects~\cite{calli2015ycb}, the reconstructions deviate from the scans by only about 1\,mm on average and yield closely matching FoundationPose trajectories, so we use the reconstructions throughout.

\subsection{Ground Truth and Metrics}\label{sec:protocol}

To validate the accuracy of our framework, we evaluate object-pose accuracy, using the motion-capture results as our ground truth (GT). The motion-capture and camera coordinate frames were aligned and the offsets of the motion-capture markers with respect to the mesh were calibrated through static scenes.

All trackers produce predictions on the same 47,222 frames in which at least four motion-capture markers are visible with a fit residual within 2\,mm. To stratify them by occlusion we evaluate (\ref{eq:vis}) at the GT pose, using the full mesh and the unmasked depth, and call the result $v_d^{\mathrm{GT}}$: a frame is occluded when $v_d^{\mathrm{GT}}<0.55$ and visible otherwise. The threshold was set from the first FoundationPose failure per session. This threshold is used only to stratify the evaluation and does not affect VICON tracking or optimization. This leaves 24,285 occluded and 22,937 visible frames.

\begin{figure}[t]
\centering
\includegraphics[width=\columnwidth]{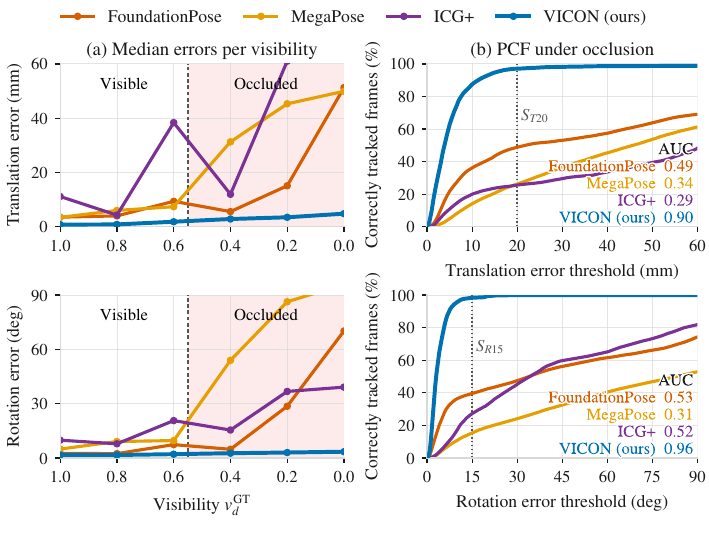}
\caption{Object-pose errors over the 40 motion-capture sessions (top: translation; bottom: symmetry-aware rotation). (a)~Median error of each tracker binned by the visibility $v_d^{\mathrm{GT}}$; shading marks the occluded range. (b)~PCF on the 24,285 occluded frames as a function of the error threshold; dotted lines mark the 20\,mm and 15$^\circ$ thresholds used for $S_{T20}$ and $S_{R15}$, and the numbers give the AUC. Medians are used in (a) because diverging baseline errors would dominate a mean.}
\label{fig:occlusion}
\end{figure}

Accuracy is measured as the translation error $e_{T,k}=10^3\|\pos_k-\pos_k^*\|_2$ (mm) and the symmetry-aware rotation error with respect to the GT pose $(\R_k^*,\pos_k^*)$,
\begin{equation}
e_{R,k}=\frac{180}{\pi}\min_{\mathbf S\in\mathcal S_O}\cos^{-1}\frac{\operatorname{tr}[(\R_k^*\mathbf S)^T\R_k]-1}{2},\label{eq:rotation-error}
\end{equation}
where the argument of $\cos^{-1}$ is clipped to $[-1,1]$ and $\mathcal S_O$ holds the rotations that leave the object mesh unchanged: $90^\circ$ steps about the mesh $z$-axis for the block, $180^\circ$ for the phone (flips excluded), and only $\{I\}$ for the rest. We report the percentage of correctly tracked frames, $\mathrm{PCF}_m(\tau;\mathcal K)$, i.e.\ the fraction (in \%) of frames in $\mathcal K$ with $e_{m,k}\leq\tau$ for $m\in\{T,R\}$, and abbreviate $S_{T20}=\mathrm{PCF}_T(20\,\mathrm{mm})$ and $S_{R15}=\mathrm{PCF}_R(15^\circ)$. A frame \emph{fails} when $e_{T,k}>20$\,mm or $e_{R,k}>15^\circ$, since a wrong orientation spoils the contact geometry even at the right position; the tolerances are commensurate because a $15^\circ$ rotation displaces a surface point on these objects by roughly 20\,mm. Failed-frame rates are taken over all 47,222 frames, and Fig.~\ref{fig:occlusion}(b) also reports the normalized area under the PCF curve up to $\tau_{\max}=60$\,mm or 90$^\circ$.

\begin{figure*}[t]
\centering
\includegraphics[width=0.9\textwidth]{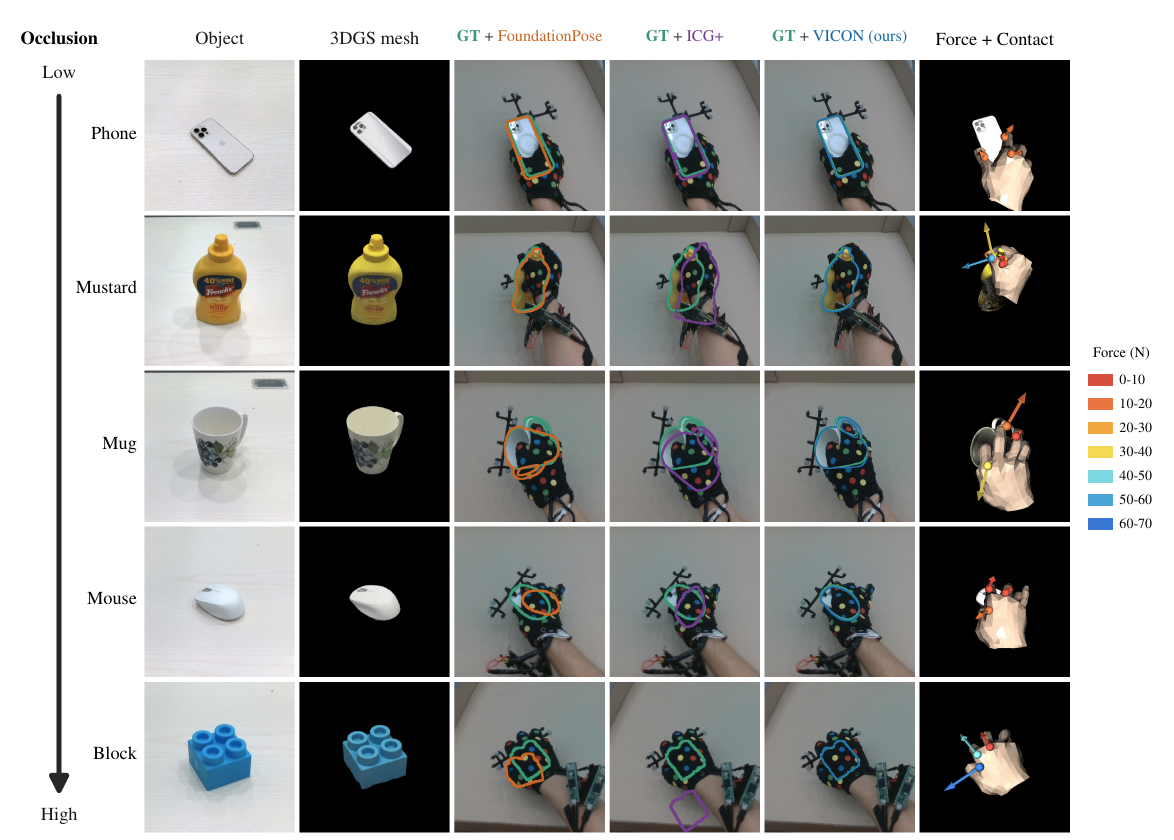}
\caption{Qualitative results with increasing hand occlusion from top to bottom. Columns: object, 3DGS mesh, and the input image overlaid with the ground-truth pose (green) and the pose of FoundationPose, ICG+, and VICON. Last column: rotated view of the tracked hand, visualized as a MANO~\cite{romero2017mano} mesh, and the VICON object pose with the contact points and force arrows of the recognized FSRs, colored by the calibrated normal force.}
\label{fig:qual}
\end{figure*}

\subsection{Comparison with Baseline Trackers}\label{sec:baselines}

Our baselines are FoundationPose~\cite{wen2024foundationpose}, MegaPose~\cite{labbe2022megapose} run sequentially on RGB-D, and ICG+~\cite{stoiber2023icgplus} with its region, depth, and ORB modalities in its YCB-Video configuration. All of them saw the same RGB-D frames and meshes, started from the first FoundationPose pose registered with a text-prompted SAM~3 mask~\cite{carion2026sam3}, and then tracked all 40 sessions without external pose feedback or reset from the GT. None of the baselines was given FSR or glove measurements, and Table~\ref{tab:baselines} compares the proposed VICON and baselines.

\begin{table}[!t]
\caption{Comparison with baseline trackers}
\label{tab:baselines}
\centering
\footnotesize
\setlength{\tabcolsep}{1.4pt}
\renewcommand{\arraystretch}{1.1}
\begin{tabular*}{\columnwidth}{@{\extracolsep{\fill}}lccccccccc@{}}
\toprule
 & Failed & \multicolumn{4}{c}{Success rate} & \multicolumn{4}{c}{Median error} \\
\cmidrule(lr){3-6}\cmidrule(lr){7-10}
 & & \multicolumn{2}{c}{Visible} & \multicolumn{2}{c}{Occluded}
   & \multicolumn{2}{c}{Visible} & \multicolumn{2}{c}{Occluded} \\
\cmidrule(lr){3-4}\cmidrule(lr){5-6}\cmidrule(lr){7-8}\cmidrule(lr){9-10}
Method & frames & $S_{T20}$ & $S_{R15}$ & $S_{T20}$ & $S_{R15}$
       & $T$ & $R$ & $T$ & $R$ \\[-1.5pt]
 & (\%) & (\%) & (\%) & (\%) & (\%) & (mm) & (deg) & (mm) & (deg) \\
\midrule
FoundationPose & 50.9 & 63.6 & 62.5 & 48.8 & 39.6 & 4.0 & 2.7 & 21.8 & 33.9 \\
MegaPose & 62.6 & 79.3 & 64.5 & 26.2 & 15.3 & 4.4 & 5.9 & 45.7 & 82.9 \\
ICG+ & 64.6 & 52.3 & 57.9 & 25.6 & 27.3 & 6.9 & 9.5 & 61.1 & 34.3 \\
VICON (ours) & \textbf{2.5} & \textbf{100.0} & \textbf{99.6} & \textbf{97.1} & \textbf{98.4}
             & \textbf{0.9} & \textbf{1.7} & \textbf{3.9} & \textbf{3.0} \\
\bottomrule
\end{tabular*}
\par\vspace{4pt}
\begin{minipage}{\columnwidth}\fontsize{8}{9}\selectfont\raggedright
Failed frames: $T>20$\,mm or $R>15^\circ$ over all 47,222 frames (Sec.~\ref{sec:protocol}). Visible/occluded: $v_d^{\mathrm{GT}}\geq0.55$ / $<0.55$ based on the threshold set from the first FoundationPose failure per session (Sec.~\ref{sec:protocol}). $S_{T20}$/$S_{R15}$: PCF (Sec.~\ref{sec:protocol}) at 20\,mm/15$^\circ$. $R$: symmetry-aware rotation error (\ref{eq:rotation-error}). Bold: best per column.
\end{minipage}
\vspace{-4pt}
\end{table}

VICON reduces the failed-frame rate to 2.5\%, compared with 50.9--64.6\% for the baselines (Table~\ref{tab:baselines}), with occluded median errors of 3.9\,mm and 3.0$^\circ$ (Fig.~\ref{fig:occlusion}). FoundationPose fails on 38.8\% of the visible frames; all but two frames follow an occluded failure without recovery, so the errors persist once visibility returns. Sequential tracking refines from the previous frame without re-checking the estimate against the observation, whereas VICON down-weights degraded poses, carries the object through the grasp interval with the hand-motion prior, and re-measures from there. VICON has the lowest occluded median translation error in 36 of the 40 sessions, with 23 sessions containing no failures, and the per-object occluded medians range from 2.3 to 5.1\,mm. Fig.~\ref{fig:qual} illustrates tracking and contact annotations under increasing hand occlusion, where FoundationPose and ICG+ deviate from the GT as the hand covers the object while VICON remains close.

\begin{table}[t]
\caption{Component ablations}
\label{tab:design}
\centering
\footnotesize
\setlength{\tabcolsep}{1.4pt}
\renewcommand{\arraystretch}{1.1}
\begin{tabular*}{\columnwidth}{@{\extracolsep{\fill}}lccccccccc@{}}
\toprule
 & Failed & \multicolumn{4}{c}{Success rate} & \multicolumn{4}{c}{Median error} \\
\cmidrule(lr){3-6}\cmidrule(lr){7-10}
 & & \multicolumn{2}{c}{Visible} & \multicolumn{2}{c}{Occluded}
   & \multicolumn{2}{c}{Visible} & \multicolumn{2}{c}{Occluded} \\
\cmidrule(lr){3-4}\cmidrule(lr){5-6}\cmidrule(lr){7-8}\cmidrule(lr){9-10}
Variant & frames & $S_{T20}$ & $S_{R15}$ & $S_{T20}$ & $S_{R15}$
        & $T$ & $R$ & $T$ & $R$ \\[-1.5pt]
 & (\%) & (\%) & (\%) & (\%) & (\%) & (mm) & (deg) & (mm) & (deg) \\
\midrule
w/o vis.\ weighting & 28.9 & 93.7 & 84.8 & 74.1 & 63.3 & 3.0 & 1.9 & 5.9 & 6.3 \\
w/o hand motion & 18.1 & 99.9 & 97.5 & 73.7 & 78.4 & 1.0 & 1.8 & 5.2 & 5.2 \\
w/o contact timing & 23.9 & 99.7 & 92.9 & 85.4 & 65.8 & 1.4 & 1.9 & 7.6 & 9.4 \\
VICON (full) & \textbf{2.5} & \textbf{100.0} & \textbf{99.6} & \textbf{97.1} & \textbf{98.4}
             & \textbf{0.9} & \textbf{1.7} & \textbf{3.9} & \textbf{3.0} \\
\bottomrule
\end{tabular*}
\par\vspace{4pt}
\begin{minipage}{\columnwidth}\fontsize{8}{9}\selectfont\raggedright
Each row removes one component from the final refinement, with the contact states of Sec.~\ref{sec:contact} held fixed. Columns as in Table~\ref{tab:baselines}. Bold: best per column.
\end{minipage}
\vspace{-4pt}
\end{table}

\subsection{Component Ablations}\label{sec:ablation}

Table~\ref{tab:design} ablates the second (final) refinement on the same frames, with the hand poses and the contact states of Sec.~\ref{sec:contact} fixed, so the variants differ only in how the refinement uses the visual and contact evidence. Each variant removes one component and retains the other two.

\textit{Visibility weighting.} The acceptance gates, hand mask, and outlier checks are removed and the observation deviations fixed to their $v_d=0.5$ values, so unreliable FoundationPose candidates enter the graph at that fixed weight.

\textit{Hand-motion prior.} We disable hand-motion initialization, propagation, the contact-gated prior, the grasp links ($\mathcal A$, $\mathcal G$), and the synthetic poses, leaving occluded frames dependent on visual evidence that is itself degraded. The occluded-frame medians include stationary periods and therefore do not isolate the contribution during transport.

\textit{Tactile contact timing.} The contact gate is disabled, so the hand-motion prior is applied to every usable hand increment, including intervals in which the hand moves without the object.

Removing any component increases the failed-frame rate from 2.5\% to 18.1--28.9\%. Paired comparisons of the per-session occluded median translation errors favor full VICON over each ablation (40~sessions; two-sided sign tests, Holm-corrected $p < 0.0005$).

\subsection{Dataset and Contact-Force Annotations}\label{sec:dataset}
The current dataset consists of all 71 sessions (including 40 motion-capture sessions): 81,307 RGB-D frames, meshes, hand skeleton trajectories and refined object trajectories, and the contact records of Sec.~\ref{sec:proximity}. The calibrated normal forces have a median of 10.9\,N and a 90th percentile of 27.0\,N; 83\% of them fall between 0 and 20\,N, the range for which the calibration error is reported in Sec.~\ref{sec:proximity}.

\section{Conclusion}\label{sec:conclusion}
We present VICON, a visual-inertial-contact based hand-object tracking framework that is robust to occlusion and enables scalable collection of contact-rich demonstrations. Using a single RGB-D camera and the VICON-glove, our system synchronously captures hand-object motion, contact points, and contact forces. Furthermore, our offline refinement recovers robust object trajectories even under severe occlusion by weighting visual estimates by visibility and admitting hand motion constraints only under measured contact. Using our framework, the failed-frame rate falls from 50.9\% (FoundationPose) to 2.5\%. The current dataset covers single-hand sequences with five rigid objects. Future work will add subjects, object classes, and faster motions, while relaxing the fixed-grasp prior to support full in-hand manipulation, and evaluate these synchronized kinematic and tactile records in policy learning.

\bibliographystyle{IEEEtran}
\bibliography{refs}

@IEEEtranBSTCTL{IEEEexample:BSTcontrol,
      CTLuse_forced_etal        = "yes",
      CTLmax_names_forced_etal  = "6",
      CTLnames_show_etal        = "1"
    }

@inproceedings{tao2025dexwild,
      author = {Tao, Tony and Srirama, Mohan Kumar and Liu, Jason Jingzhou and Shaw, Kenneth and Pathak, Deepak},
      title = {{DexWild}: Dexterous Human Interactions for In-the-Wild Robot Policies},
      booktitle = {Proc. Robot.: Sci. Syst. (RSS)},
      year = {2025},
      doi = {10.15607/RSS.2025.XXI.075}
    }

@inproceedings{park2022handoccnet,
      author = {Park, JoonKyu and Oh, Yeonguk and Moon, Gyeongsik and Choi, Hongsuk and Lee, Kyoung Mu},
      title = {{HandOccNet}: Occlusion-Robust {3D} Hand Mesh Estimation Network},
      booktitle = {Proc. IEEE/CVF Conf. Comput. Vis. Pattern Recognit. (CVPR)},
      year = {2022},
      pages = {1486--1495},
      doi = {10.1109/CVPR52688.2022.00155}
    }

@inproceedings{qi2024hoisdf,
      author = {Qi, Haozhe and Zhao, Chen and Salzmann, Mathieu and Mathis, Alexander},
      title = {{HOISDF}: Constraining {3D} Hand-Object Pose Estimation with Global Signed Distance Fields},
      booktitle = {Proc. IEEE/CVF Conf. Comput. Vis. Pattern Recognit. (CVPR)},
      year = {2024},
      pages = {10392--10402},
      doi = {10.1109/CVPR52733.2024.00989}
    }

@inproceedings{fan2024hold,
      author = {Fan, Zicong and Parelli, Maria and Kadoglou, Maria Eleni and Chen, Xu and Kocabas, Muhammed and Black, Michael J. and Hilliges, Otmar},
      title = {{HOLD}: Category-agnostic {3D} Reconstruction of Interacting Hands and Objects from Video},
      booktitle = {Proc. IEEE/CVF Conf. Comput. Vis. Pattern Recognit. (CVPR)},
      year = {2024},
      pages = {494--504},
      doi = {10.1109/CVPR52733.2024.00054}
    }

@article{lee2021vist,
      author = {Lee, Yongseok and Do, Wonkyung and Yoon, Hanbyeol and Heo, Jinuk and Lee, WonHa and Lee, Dongjun},
      title = {Visual-inertial hand motion tracking with robustness against occlusion, interference, and contact},
      journal = {Sci. Robot.},
      volume = {6},
      number = {58},
      pages = {eabe1315},
      year = {2021},
      doi = {10.1126/scirobotics.abe1315}
    }

@inproceedings{choi2026robusthand,
      author = {Choi, Hyelim and Park, Hyunreal and Ji, Harim and Lee, Somang and Lee, Youngseon and Lee, Yongseok and Lee, Dongjun},
      title = {Robust Hand Tracking from Visual-Inertial Fusion},
      booktitle = {Proc. IEEE Int. Conf. Robot. Autom. (ICRA)},
      year = {2026}
    }

@inproceedings{brahmbhatt2020contactpose,
      author = {Brahmbhatt, Samarth and Tang, Chengcheng and Twigg, Christopher D. and Kemp, Charles C. and Hays, James},
      title = {{ContactPose}: A Dataset of Grasps with Object Contact and Hand Pose},
      booktitle = {Proc. Eur. Conf. Comput. Vis. (ECCV)},
      year = {2020},
      pages = {361--378},
      doi = {10.1007/978-3-030-58601-0_22}
    }

@inproceedings{fan2023arctic,
      author = {Fan, Zicong and Taheri, Omid and Tzionas, Dimitrios and Kocabas, Muhammed and Kaufmann, Manuel and Black, Michael J. and Hilliges, Otmar},
      title = {{ARCTIC}: A Dataset for Dexterous Bimanual Hand-Object Manipulation},
      booktitle = {Proc. IEEE/CVF Conf. Comput. Vis. Pattern Recognit. (CVPR)},
      year = {2023},
      pages = {12943--12954},
      doi = {10.1109/CVPR52729.2023.01244}
    }

@inproceedings{hoque2026egodex,
      author    = {Hoque, Ryan and Huang, Peide and Yoon, David J. and Sivapurapu, Mouli and Zhang, Jian},
      title     = {{EgoDex}: Learning Dexterous Manipulation from Large-Scale Egocentric Video},
      booktitle = {Proc. Int. Conf. Learn. Represent. (ICLR)},
      year      = {2026}
    }

@article{song2025opentouch,
      author = {Song, Yuxin Ray and Li, Jinzhou and Fu, Rao and Murphy, Devin and Zhou, Kaichen and Shiv, Rishi and Li, Yaqi and Xiong, Haoyu and Owens, Crystal Elaine and Du, Yilun and Luo, Yiyue and Cheng, Xianyi and Torralba, Antonio and Matusik, Wojciech and Liang, Paul Pu},
      title = {{OpenTouch}: Bringing Full-Hand Touch to Real-World Interaction},
      journal = {arXiv preprint arXiv:2512.16842},
      year = {2025}
    }

@inproceedings{yu2025forcevla,
      author = {Yu, Jiawen and Liu, Hairuo and Yu, Qiaojun and Ren, Jieji and Hao, Ce and Ding, Haitong and Huang, Guangyu and Huang, Guofan and Song, Yan and Cai, Panpan and Zhang, Wenqiang and Lu, Cewu},
      title = {{ForceVLA}: Enhancing {VLA} Models with a Force-aware {MoE} for Contact-rich Manipulation},
      booktitle = {Adv. Neural Inf. Process. Syst. (NeurIPS)},
      volume = {38},
      year = {2025},
      pages = {103609--103639},
      doi = {10.52202/085713-3124}
    }

@article{chen2025dexforce,
      author = {Chen, Claire and Yu, Zhongchun and Choi, Hojung and Cutkosky, Mark and Bohg, Jeannette},
      title = {{DexForce}: Extracting Force-Informed Actions from Kinesthetic Demonstrations for Dexterous Manipulation},
      journal = {IEEE Robot. Autom. Lett.},
      volume = {10},
      number = {6},
      pages = {6416--6423},
      year = {2025},
      doi = {10.1109/LRA.2025.3568318}
    }

@article{ni2026tactidex,
      author = {Ni, Suting and Zhang, Hanbing and Wei, Zhenyu and Chen, Guo and Zhang, Chixuan and Shi, Ye and Wang, Jingya},
      title = {{TactiDex}: A Real-World Tactile-Guided Benchmark for Human-Like Dexterous Manipulation},
      journal = {arXiv preprint arXiv:2607.09190},
      year = {2026}
    }

@inproceedings{labbe2022megapose,
      author    = {Labb{\'e}, Yann and Manuelli, Lucas and Mousavian, Arsalan and Tyree, Stephen and Birchfield, Stan and Tremblay, Jonathan and Carpentier, Justin and Aubry, Mathieu and Fox, Dieter and Sivic, Josef},
      title     = {{MegaPose}: {6D} Pose Estimation of Novel Objects via Render \& Compare},
      booktitle = {Proc. Conf. Robot Learn. (CoRL)},
      series    = {Proc. Mach. Learn. Res.},
      volume    = {205},
      year      = {2023},
      pages     = {715--725}
    }

@article{romero2017mano,
      author = {Romero, Javier and Tzionas, Dimitrios and Black, Michael J.},
      title = {Embodied Hands: Modeling and Capturing Hands and Bodies Together},
      journal = {ACM Trans. Graph.},
      volume = {36},
      number = {6},
      pages = {245:1--245:17},
      year = {2017},
      doi = {10.1145/3130800.3130883}
    }

@inproceedings{wen2024foundationpose,
      author = {Wen, Bowen and Yang, Wei and Kautz, Jan and Birchfield, Stan},
      title = {{FoundationPose}: Unified {6D} Pose Estimation and Tracking of Novel Objects},
      booktitle = {Proc. IEEE/CVF Conf. Comput. Vis. Pattern Recognit. (CVPR)},
      year = {2024},
      pages = {17868--17879},
      doi = {10.1109/CVPR52733.2024.01692}
    }

@article{shin2026compose,
      author  = {Shin, Jisu and Lee, Junoh and Lee, JunGyu and Bae, Inhwan and Lee, Dohyeon and Im, Hokyun and Lee, Youngwoon and Jeon, Hae-Gon},
      title   = {{ComPose}: When to Trust Hands for Object Pose Tracking},
      journal = {arXiv preprint arXiv:2605.23523},
      year    = {2026}
    }

@article{sundaram2019stag,
      author = {Sundaram, Subramanian and Kellnhofer, Petr and Li, Yunzhu and Zhu, Jun-Yan and Torralba, Antonio and Matusik, Wojciech},
      title = {Learning the Signatures of the Human Grasp Using a Scalable Tactile Glove},
      journal = {Nature},
      volume = {569},
      number = {7758},
      pages = {698--702},
      year = {2019},
      doi = {10.1038/s41586-019-1234-z}
    }

@article{gonzalez2014hand,
      author = {Gonzalez, Franck and Gosselin, Florian and Bachta, Wael},
      title = {Analysis of Hand Contact Areas and Interaction Capabilities During Manipulation and Exploration},
      journal = {IEEE Trans. Haptics},
      volume = {7},
      number = {4},
      pages = {415--429},
      year = {2014},
      doi = {10.1109/TOH.2014.2321395}
    }

@inproceedings{wen2026fastfoundationstereo,
      author    = {Wen, Bowen and Dewan, Shaurya and Birchfield, Stan},
      title     = {{Fast-FoundationStereo}: Real-Time Zero-Shot Stereo Matching},
      booktitle = {Proc. IEEE/CVF Conf. Comput. Vis. Pattern Recognit. (CVPR)},
      year      = {2026},
      pages     = {7513--7524}
    }

@inproceedings{carion2026sam3,
      author    = {Carion, Nicolas and Gustafson, Laura and Hu, Yuan-Ting and Debnath, Shoubhik and Hu, Ronghang and Suris, Didac and Ryali, Chaitanya and Alwala, Kalyan Vasudev and Khedr, Haitham and Huang, Andrew and others},
      title     = {{SAM} 3: Segment Anything with Concepts},
      booktitle = {Proc. Int. Conf. Learn. Represent. (ICLR)},
      year      = {2026}
    }

@article{kerbl2023gs,
      author  = {Kerbl, Bernhard and Kopanas, Georgios and Leimk{\"u}hler, Thomas and Drettakis, George},
      title   = {{3D} {G}aussian Splatting for Real-Time Radiance Field Rendering},
      journal = {ACM Trans. Graph.},
      volume  = {42},
      number  = {4},
      pages   = {139:1--139:14},
      year    = {2023},
      doi     = {10.1145/3592433}
    }

@inproceedings{calli2015ycb,
      author = {Calli, Berk and Singh, Arjun and Walsman, Aaron and Srinivasa, Siddhartha and Abbeel, Pieter and Dollar, Aaron M.},
      title = {The {YCB} Object and Model Set: Towards Common Benchmarks for Manipulation Research},
      booktitle = {Proc. Int. Conf. Adv. Robot. (ICAR)},
      year = {2015},
      pages = {510--517},
      doi = {10.1109/ICAR.2015.7251504}
    }

@inproceedings{guedon2024sugar,
      author    = {Gu{\'e}don, Antoine and Lepetit, Vincent},
      title     = {{SuGaR}: Surface-Aligned {G}aussian Splatting for Efficient {3D} Mesh Reconstruction and High-Quality Mesh Rendering},
      booktitle = {Proc. IEEE/CVF Conf. Comput. Vis. Pattern Recognit. (CVPR)},
      year      = {2024},
      pages     = {5354--5363},
      doi       = {10.1109/CVPR52733.2024.00512}
    }

@inproceedings{schonberger2016sfm,
      author    = {Sch{\"o}nberger, Johannes Lutz and Frahm, Jan-Michael},
      title     = {Structure-from-Motion Revisited},
      booktitle = {Proc. IEEE Conf. Comput. Vis. Pattern Recognit. (CVPR)},
      year      = {2016},
      pages     = {4104--4113},
      doi       = {10.1109/CVPR.2016.445}
    }

@inproceedings{sridhar2016joint,
      author    = {Sridhar, Srinath and Mueller, Franziska and Zollh{\"o}fer, Michael and Casas, Dan and Oulasvirta, Antti and Theobalt, Christian},
      title     = {Real-time Joint Tracking of a Hand Manipulating an Object from {RGB-D} Input},
      booktitle = {Proc. Eur. Conf. Comput. Vis. (ECCV)},
      year      = {2016},
      pages     = {294--310},
      doi       = {10.1007/978-3-319-46475-6_19}
    }

@inproceedings{stoiber2023icgplus,
      author    = {Stoiber, Manuel and Elsayed, Mariam and Reichert, Anne E. and Steidle, Florian and Lee, Dongheui and Triebel, Rudolph},
      title     = {Fusing Visual Appearance and Geometry for Multi-Modality {6DoF} Object Tracking},
      booktitle = {Proc. IEEE/RSJ Int. Conf. Intell. Robots Syst. (IROS)},
      year      = {2023},
      pages     = {1170--1177},
      doi       = {10.1109/IROS55552.2023.10341961}
    }

@misc{interlink_fsr400,
      author       = {{Interlink Electronics}},
      title        = {{FSR} 400 Series Data Sheet},
      howpublished = {\url{https://www.interlinkelectronics.com/downloads/datasheets/fsr-400-series-datasheet.pdf}},
      note         = {{P/N} {PDS-10004-C}, accessed Sep. 12, 2026}
    }

@inproceedings{li2025fsglove,
      author    = {Li, Yutong and Zhang, Jieyi and Xu, Wenqiang and Tang, Tutian and Lu, Cewu},
      title     = {{FSGlove}: An Inertial-Based Hand Tracking System with Shape-Aware Calibration},
      booktitle = {Proc. IEEE/RSJ Int. Conf. Intell. Robots Syst. (IROS)},
      year      = {2025},
      pages     = {18304--18311},
      doi       = {10.1109/IROS60139.2025.11245803}
    }

\end{document}